\documentclass[review]{elsarticle}
\makeatletter
\def\ps@pprintTitle{%
 \let\@oddhead\@empty
 \let\@evenhead\@empty
 \def\@oddfoot{\centerline{\thepage}}%
 \let\@evenfoot\@oddfoot}
\makeatother
\usepackage{hyperref}
\usepackage{soul}
\usepackage{xcolor}

\newcommand{\unsupervisedModule}{Persian Syntactic Information Extractor}
\newcommand{\unsupervisedModulShorten}{PSIE}

\begin{document}
\begin{frontmatter}

\title{FarsBase-KBP: A Knowledge Base Population System for the Persian Knowledge Graph}

\author[mymainaddress]{Majid Asgari-Bidhendi}
\ead{majid.asgari@gmail.com}

\author[mymainaddress]{Behrooz Janfada}
\ead{behrooz.janfada@gmail.com}

\author[mymainaddress]{Behrouz Minaei-Bidgoli\corref{mycorrespondingauthor}}
\cortext[mycorrespondingauthor]{Corresponding author}
\ead{b\_minaei@iust.ac.ir}

\address[mymainaddress]{Computer Engineering School, Iran University of Science and Technology, Tehran, Iran}

%\maketitle

\begin{abstract}
While most of the knowledge bases already support the English language, there is only one knowledge base for the Persian language, known as FarsBase, which is automatically created via semi-structured web information. Unlike English knowledge bases such as Wikidata, which have tremendous community support, the population of a knowledge base like FarsBase must rely on automatically extracted knowledge. Knowledge base population can let FarsBase keep growing in size, as the system continues working. In this paper, we present a knowledge base population system for the Persian language, which extracts knowledge from unlabeled raw text, crawled from the Web. The proposed system consists of a set of state-of-the-art modules such as an entity linking module as well as information and relation extraction modules designed for FarsBase. Moreover, a canonicalization system is introduced to link extracted relations to FarsBase properties. Then, the system uses knowledge fusion techniques with minimal intervention of human experts to integrate and filter the proper knowledge instances, extracted by each module.
To evaluate the performance of the presented knowledge base population system, we present the first gold dataset for benchmarking knowledge base population in the Persian language, which consisting of 22015 FarsBase triples and verified by human experts. The evaluation results demonstrate the efficiency of the proposed system.
\end{abstract}

\begin{keyword}
Knowledge Extraction\sep Knowledge Graph\sep Canonicalization\sep Natural Language Processing\sep Persian Language
\end{keyword}

\end{frontmatter}

%\linenumbers

\section{Introduction}

Knowledge Base Population (KBP) is the process of populating (or building from scratch) a Knowledge Base (KB) with new knowledge elements. A considerable number of Natural Language Processing (NLP) tasks, such as question answering, use knowledge bases. Thus, there is an unfading need for more completed and populated knowledge bases.   

FarsBase \cite{asgari2019farsbase} is the first knowledge base in the Persian language. Despite other knowledge bases like DBpedia~\cite{auer2007dbpedia} and BabelNet~\cite{navigli2010babelnet}, which minimally support the Persian language, FarsBase is specially designed for the Persian Language. Similar to other knowledge bases, FarsBase faces some severe challenges such as staying up-to-date and expanding with existing knowledge. Some knowledge bases such as Wikidata~\cite{vrandevcic2014wikidata} rely on the human resources to annotate structured data and to prevent the entrance of erroneous knowledge instances to the knowledge base. Unlike Wikidata, FarsBase does not have such community support, which emphasized the need for a system to extract knowledge automatically and to prevent wrong relation instances from being passed to the knowledge base. In this paper, we introduce FarsBase Knowledge Base Population System (FarsBase-KBP) to address this issue.

Note that, while our KBP system uses different modules employing different methods for relation extraction, the extracted knowledge by them should be checked to detect possible redundancy or conflict. For example, if two modules extract two different birthdays for the same person, it means that one or both of these modules have generated erroneous fact, indicating that there is a conflict that should be detected. Additionally, there is also the problem of mapping predicates, subjects, and objects in the extracted triples, which can be extracted as raw text, to the canonicalized predicates and entities in the knowledge base. We address these issues using canonicalization techniques. It should also be noted that despite English and other high-resource languages which can rely on annotated data and use supervised methods, in Persian, which is considered a low-resource language, due to the lack of required training gold datasets, supervised methods are generally not applicable. To overcome this problem, we employed unsupervised and distantly supervised methods which do not require such data.

Our contributions are as follows:
\begin{enumerate}
    \item Unlike Relation Extraction (RE) systems, in an Information Extraction (IE) system, the relations are not canonicalized, the arguments of the extracted relations are in the form of plain text, and they are not linked to an existing knowledge base. To address this issue, we proposed an entity linking system which links subjects and objects to knowledge base entities, alongside with a canonicalization system which maps relations to knowledge (see section \ref{sec:mapper}). The entity linking system applies an entity linking method for the Persian language, namely ParsEL~\cite{asgaribidhendi2020parsel}, to link the arguments of the relation as the subject and object in a FarsBase triple. For the first time, we introduce a canonicalization system which is designed especially for the Persian language and links relationship types to the pre-defined FarsBase predicate set. In this linking process, we use current triples and mapping data in FarsBase, as well as extracted knowledge from other FarsBase-KBP modules. Both entity linking and canonicalization systems are state-of-the-art in the Persian language.
    \item To the best of our knowledge, there are few studies on relation and knowledge extraction for the Persian language, and above that, there are no studies on canonicalization (relation to knowledge mapping) and knowledge fusion in the Persian language as well. Here we propose four modules for information extraction and two modules for relation extraction; all of them are innovative and state-of-the-art methods for the Persian language. For example, the Dependency Pattern (DP) and \unsupervisedModule (\unsupervisedModulShorten{}) modules introduced and devised for the first time in FarsBase-KBP. TokensRegex is also used in the Persian language for the first time and uses FarsBase hirechial ontology classes instead NER.
    \item For the first time, we introduce and publish a gold dataset by which we evaluate the performance of knowledge extraction in FarsBase-KBP. This dataset contains 22015 sentences, in which the entities and relation types are linked to the FarsBase ontology. This gold dataset can be reused for benchmarking KBP systems in the Persian language.
\end{enumerate}

The rest of this paper is structured as follows. In section \ref{related-work}, a review of the literature of knowledge base population will be presented, then we mention other researches that employ knowledge fusion to make use of different sources for knowledge extraction, and finally, we discuss studies on knowledge extraction and knowledge base population for the Persian language. In section \ref{proposed-approach}, we give a brief background of other researches related to each of the modules of FarsBase-KBP, and then we explain how each component of our system operates and how all these components work together to improve the quality of the extracted knowledge. In the last two sections, we present the results of our experiments and the obtained conclusions.

\section{Related Work}
\label{related-work}
Knowledge Base Population is defined as the process of extending a knowledge base with information extracted from the text. The goal is to update the knowledge base or keeping it current with new information \cite{glass2018dataset}. 
In this section, we first review literature in the field of knowledge base population, then we categorize and describe studies in this field. The last subsection presents related studies focused on the Persian language.
\subsection{Automatic Knowledge Base Population}
Automatic knowledge base construction and population have recently received significant attention in academic researches. As the size of the knowledge bases kept growing, the need for automatic construction and population of knowledge bases arose. Existing knowledge bases are usually highly incomplete. For example, only 6.2\% of \textit{persons} from Freebase \cite{bollacker2008freebase} have \textit{place of birth} \cite{min2013distant}, and only 1\% of them have \textit{ethnicity} \cite{west2014knowledge}. Additionally, manual completion of existing knowledge bases is expensive and time-consuming. Therefore, automatic construction of knowledge bases from scratch, populating them with missing information, and adding new knowledge to them has attracted a lot of academic attention \cite{adel2018deep}.

Text Analysis Conference (TAC) is an annual series of open technology evaluations organized by the National Institute of Standards and Technology (NIST). The KBP track of TAC encourages the development of the systems that can extract information from unstructured text in order to populate an existing knowledge base or to construct a cold-start (built from scratch) knowledge base \cite{getman2018laying}. TAC KBP track consists of several tasks such as entity discovery and linking, and slot filling \cite{adel2018deep}. Slot Filling is a version of KBP where specific missing information (slots) are searched through the document collection and filled with desired values \cite{glass2018dataset}. 

Knowledge base population task is a follow-up task to relation extraction. KBP systems usually consist of one or more relation extractors or knowledge extractors. Relation extraction and knowledge extraction are well studied, yet growing fields of research. These extractors usually utilize basic NLP tasks such as Named Entity Recognition (NER), dependency or constituency parsing, and Entity Linking (EL) to find triples (subject, object, and predicate). Some systems such as FRED \cite{gangemi2017semantic} propose the integration of a stack of native Semantic Web (SW) machine reading tasks. FRED is a machine reader for the semantic web which extracts knowledge from the raw text in the form of RDF graph representation. This extracted knowledge can be used to populate a knowledge base \cite{consoli2015using}. We will provide more details about relation and knowledge extraction literature in the proposed approach section when FarsBase-KBP extractor modules are described.
\subsection{KBP Categories}
Regarding their method of extraction and source of information, there are four main categories of literature in this field \cite{dong2014knowledge} 

\textbf{Built on Structured Data:} Approaches which populate knowledge bases using structured data sources (like Wikipedia infoboxes), such as DBpedia \cite{auer2007dbpedia}, Freebase \cite{bollacker2008freebase} YAGO \cite{suchanek2007yago}, and YAGO2 \cite{hoffart2013yago2}.
Yago facts are automatically extracted from Wikipedia, using a combination of heuristic and rule-based methods. YAGO2 is a more recent extension of YAGO in which entities, facts, and events are anchored in both time and space.  

\textbf{Open Information Extraction, Web-scale:} Approaches which apply open Information Extraction (IE) techniques which are applied to the entire Web, ranging from Reverb~\cite{fader2011identifying}, PRISMATIC~\cite{fan2010prismatic} and OLLIE~\cite{schmitz2012open} to MINIE~\cite{gashteovski2017minie} and Graphene~\cite{cetto2018graphene}. Graphene is a recent Open IE approach that presents a two-layered hierarchical representation of syntactically simplified sentences in the form of core facts and accompanying contexts that are semantically linked by rhetorical relations.
 
\textbf{Making Taxonomies:} Approaches which construct taxonomies (is-a hierarchies), as opposed to general KBs with multiple types of predicates such as Probase \cite{wu2012probase}. Probase is a universal, general-purpose, and probabilistic taxonomy which is automatically constructed from a corpus of 1.6 billion web pages. It consists of an iterative learning algorithm to extract ``is-a'' pairs from web texts, plus a taxonomy construction algorithm to connect these pairs to a hierarchical structure.
 
\textbf{Fixed Ontology, Web-scale:} Approaches which extract information from the entire web, but use a fixed ontology (schema) such as PROSPERA~\cite{nakashole2011scalable}, DeepDive~\cite{niu2012deepdive}, Knowledge Vault~\cite{dong2014knowledge}, and Never-Ending Language Learner (NELL)~\cite{carlson2010toward}. NELL is the first knowledge base that uses automatic extraction of knowledge with very little human intervention. The original architecture of NELL consists of four knowledge extraction components. The knowledge integrator module handles knowledge fusion in NELL. This module promotes knowledge instances to beliefs if it is extracted by one high-confidence source or by multiple sources. The latest version of NELL \cite{mitchell2018never} keeps the same architecture plus five new extractor modules are added, such as an image classifier. Other more recent knowledge bases employ automatic knowledge extraction methods, such as DeepDive and Knowledge Vault. The main difference between Knowledge Vault and NELL is that Knowledge Vault fuses facts extracted from the text with the prior knowledge derived from the Freebase~\cite{bollacker2008freebase}. 

The proposed knowledge base population system, FarsBase-KBP, is classified in the last category. FarsBase-KBP knowledge fusion module utilizes a similar approach as knowledge integrator in NELL, although despite NELL, it links the entities of the extracted triples to FarsBase. FarsBase-KBP and NELL also differ in how human intervenes in the knowledge fusion module. In NELL, the triples are transferred directly to the knowledge base after the fusion stage, while at a limited daily time window, some of the knowledge base triples are examined by experts, and false triples are identified. However, in FarsBase-KBP, any triple extracted by the fusion module must be checked by a human expert and added to the knowledge base if approved.

\subsection{Entity Linking}
In the context of KBP, Entity Linking (EL) is the task of mapping all of the subjects and some of the objects of the triples in a raw text to their corresponding entities in a knowledge base. With the appearance of FarsBase, EL has become a possible task for the Persian language as well. In our previous work, we proposed ParsEL~\cite{asgaribidhendi2020parsel}, which is a language-independent end-to-end entity linking system that utilizes both context-dependent and context-independent features for entity linking. ParsEL is the first entity linking system applied to the Persian language with FarsBase as the external dataset. Our experiments showed that the proposed method outperforms Babelfy~\cite{moro2014entity} as the state-of-the-art of multilingual end-to-end entity linking algorithm.

\subsection{Information Extraction}
Information extraction (IE) is one of the essential tasks in NLP. Its purpose is to extract structured information from raw, unstructured text. An IE system receives the raw text and generates a set of triples or n-ary propositions, usually in the form of (subject, predicate, object) as structured information, in which the predicate is a part of the raw text that represents the relationship between the subject and some objects~\cite{shi2019brief}. 

Roy et al. provided a new supervised OIE approach that uses a set of unsupervised OIE systems and a small amount of tagged data. As its input features, this method utilizes the output of several unsupervised OIE systems as well as a diverse set of lexical and syntax information including word embedding, POS embedding, syntactic role embedding and dependency structure~\cite{roy2019supervising}.

\subsection{Relation Extraction}
Relation Extraction (RE) is a specific case of IE, in which entities and semantic (mapped) relations between them are identified in the input text.  In other words, an RE system can predict whether the input text has a relationship for some arguments or not. Besides, the RE system must predict which relational class from a particular ontology predicts the identified relation of the input text. Supervised RE methods require training datasets to learn the model. Generating such annotated datasets for RE is time-consuming and expensive, hence resource-poor languages lack of such datasets. In a review study, Shi et al. Showed that if it is not possible to use supervised methods for relation extraction, distant supervision methods will yield the best results~\cite{shi2019brief}. 

Trisedya et al. proposed an end-to-end RE model for KB enrichment based on a neural encoder-decoder model, utilizing distantly supervised training data with co-reference resolution and paraphrase detection~\cite{trisedya2019neural}. Gao and his colleagues have proposed Neural Snowball, a new bootstrapping method for learning new relations by transferring semantic knowledge about existing relations. They designed Relation Siamese Networks (RelSN) to learn the criteria for the similarity of relations based on their tagged data and existing relations~\cite{gao2019neural}. Smirnova and Cudré-Mauroux have reviewed RE methods utilizing distant supervision and summarized the two main challenges in this field, noisy labels automatically collected from the KB and the evaluation and training problems induced by the incompleteness of the KB~\cite{smirnova2018relation}.

\subsection{Canonicalization}
One of the essential systems which contribute to construction and population of KBs are Open Information Extraction (OIE) approaches.  However, one fundamental problem is that the relation phrases in the extracted triples of OIE system are not linked or mapped to the knowledge base ontology predicates, which leads to a large number of ambiguous and redundant triples. Canonicalization is the task of mapping the plain text relation phrases to appropriate predicates in the knowledge base. The quality of canonicalization can directly affect the quality of the KBP system.

Various methods have been proposed for canonicalization. Putri et al. proposed a novel approach based on distant supervision and a Siamese network that compares two sequences of word embeddings, representing an OIE relation and a predefined KB relation~\cite{putri2019aligning}. Vashishth et al. proposed CESI~\cite{vashishth2018cesi}, a system for Canonicalization using Embeddings and Side Information which is a novel approach that performs canonicalization above learned embeddings of Open KBs. Side information earned from AMIE~\cite{galarraga2013amie} and Stanford KBP~\cite{surdeanu2012multi}. Lin and Chen proposed an approach for canonicalization which utilizes the side information of the original data sources, including the entity linking knowledge, the types of the candidate entities detected, as well as the domain knowledge of the source text. They jointly modelled the canonicalization problem of entity and relation phrases and proposed a clustering method and demonstrated the effectiveness of this approach through extensive experiments on two different datasets~\cite{lin2019canonicalization}. Other studies have suggested clustering methods for canonicalization. Galárraga and Heitz presented a machine-learning-based strategy utilizing the AMIE algorithm~\cite{galarraga2013amie} that can canonicalize Open IE triples, by clustering synonymous names and phrases~\cite{galarraga2014canonicalizing}.

\subsection{Literature in Low-Resource Languages}
In this subsection, by low-resource language (resource-poor language), we mean a language without a lot of labelled datasets and corpora dedicated for the training of the supervised methods. If we look at KBP as a whole system, there are different approaches proposed, most of which in the TAC-KBP challenges, which typically utilizes components such as IE, RE, EL, Canonicalization and Fusion components. The TAC-KBP challenges benchmarking are performed over three languages, namely English, Spanish and Chinees, which are not low-resource languages~\cite{getman2018laying}. The suggested state-of-the-art KBP benchmarking methods such as KnowledgeNet~\cite{mesquita2019knowledgenet} and CC-DBP~\cite{glass2018dataset} is also focused on English. To the best of our knowledge, considering KBP as a whole system, no specific research has been found for low-resource languages.

However, given the subsystems of a KBP system such as IE and RE, some research has been done in low-resource languages, but most of them are cross-lingual. A few of these types of research, which are novel, up to date and state-of-the-art, are discussed subsequently. Taghizadeh et al. proposed a system for RE in Arabic by a cross-language learning method, utilizing the training data of other languages such as Universal Dependency (UD)~\cite{straka2015parsing} parsing and the similarity of UD trees in different languages and trains a RE model for Arabic text~\cite{taghizadeh2018cross}. Zakria et al. proposed a method for RE in Arabic exploiting Arabic Wikipedia articles properties. The proposed system extracts sentences that contain principle entity, secondary entity and relation from Wikipedia article, then utilizes WordNet and DBpedia to build the training set. Then a Naive Bayes Classifier is used to train and test the datasets~\cite{zakria2019relation}. AlArfaj have reviewed the different state-of-the-art RE methods for Arabic and showed that the majority of RE approaches utilize a combination of statistical and linguistic-pattern techniques. The review proposes that linguistic-pattern methods provide high precision, but their recall is very low, and patterns are specified in the regular expression form, which is challenging to cope with language variety~\cite{alarfaj2019towards}. Sarhan et al. proposed a semisupervised pattern-based bootstrapping technique for Arabic RE utilizing a dependency parser to omit negative relations as well as features like stemming, semantic expansion using synonyms, and an automatic scoring technique to measure the reliability of the generated patterns and extracted relations~\cite{sarhan2016semi}.

SUN et al. proposed a distantly supervised RE model based on Piecewise Convolutional Neural Network (PCNN) to expand the Tibetan corpus. They also added self-attention mechanism and soft-label method to decrease wrong labels, and use Embeddings from ELMo language model~\cite{peters2018deep} to solve the semantic ambiguity problem~\cite{sun2019improved}. Lkhagvasuren and Rentsendorj presented MongoIE, which is an OIE system for the Mongolian language, utilizing rule-based and classification approaches. Their classification method showed better results than the rule-based approach~cite{lkhagvasuren2020open}. Peng~\cite{peng2017jointly} suggested that learning under low-resource conditions needs special techniques when there is inadequate training data, including distant supervision, multi-domain learning, and multi-task learning. However, no particular approach proposed for low-resource languages.

\subsection{Literature in the Persian Language}
There are few studies on knowledge extraction and knowledge base population in the Persian language. Moradi et al. \cite{moradi2015commonsense} propose a system focused on certain generic relation types, such as is-a relationship, to extract relation instances. Hasti Project proposed by Shamsfard and Barforoush \cite{shamsfard2004learning} is another research which introduced an automatic ontology building approach. It extracts lexical and ontological knowledge from Persian raw text and uses logical, template-driven, and semantic analysis methods.  Momtazi and Moradiannasab \cite{momtazi2019statistical} also proposed a statistical n-gram method which extracts knowledge from unstructured Persian language texts.

\section{Proposed Approach}
\label{proposed-approach}
In this section, we describe the architecture of FarsBase-KBP and propose our approach in which extracted triples from different extractors are integrated and stored into a knowledge graph. Six extractor components are used in total, including four information extractors and two relation extractors. The predicates in extracted triples by information extractors are plain texts, not predefined relations. Consequently, their predicates are not linked to relations in the knowledge base. Thus a canonicalization process is needed to link predicates of extracted relations to the Knowledge Graph (KG) resources.

\subsection{FarsBase-KBP Architecture}
\begin{figure}
	\includegraphics[width=\linewidth]{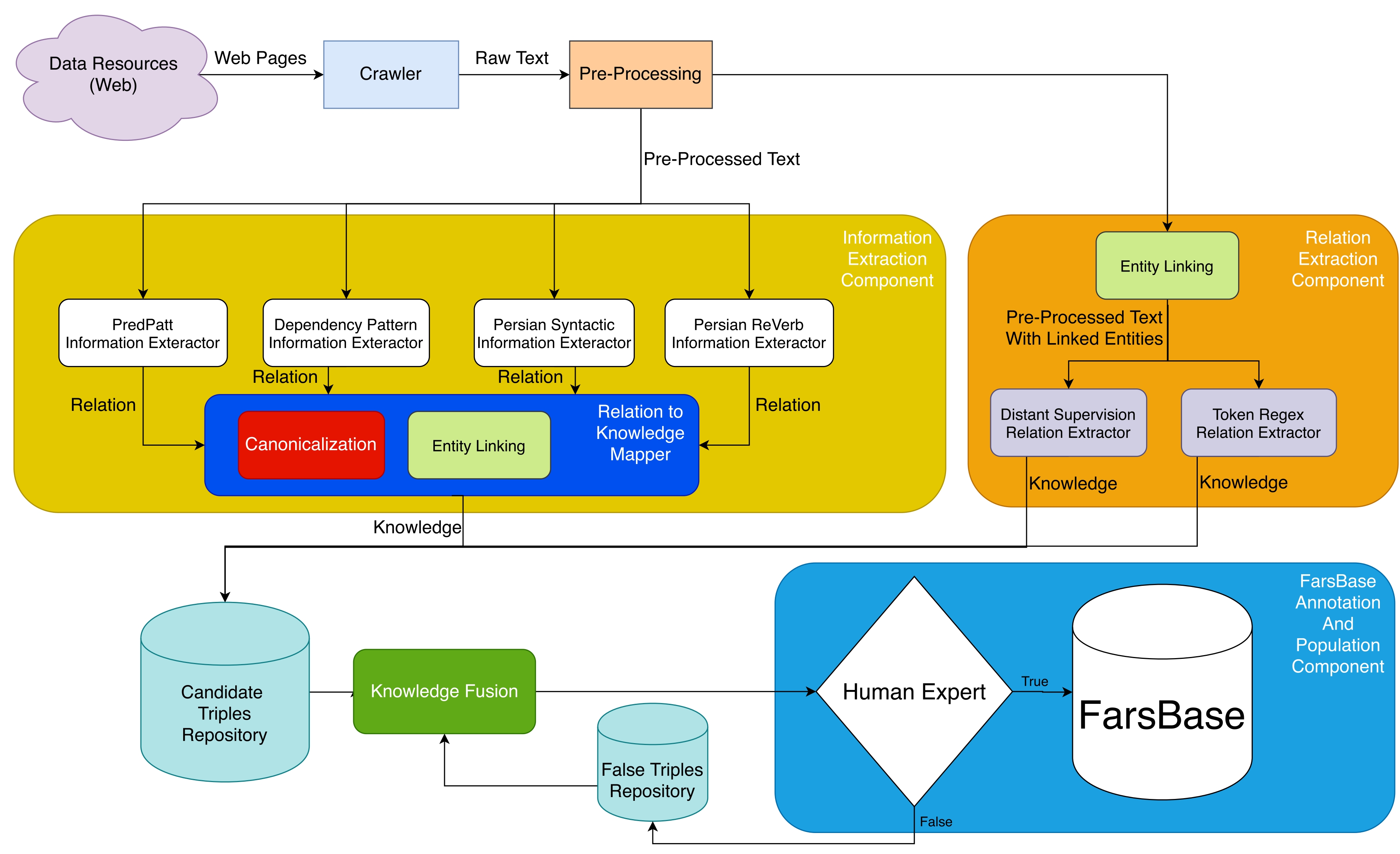}
	\caption{The architecture of FarsBase knowledge base population system (FarsBase-KBP)}
	\label{fig:Arch}
\end{figure}

The architecture of our system, FarsBase-KBP works on the top of six extractor modules and fuses their output triples. Figure \ref{fig:Arch} shows a block diagram of the architecture and data flow of FarsBase-KBP. First, a crawler crawls the Web and extracts raw texts. After applying a pre-processing stage, the extracted text will be delivered to the entity linking module. This module links entities of the pre-processed text and passes the processed text to each of the six extractor components. 
Two relation extractors produce triples which are mapped to KG entities and delivered to the Candidate-Fact-Triples-Repository (CFTR) for further process. Note that each triple has a corresponding confidence value which is assigned by its extractor module. Four information extractors work independently and produce information triples containing a subject, object, and not-canonicalized predicate. Then, these triples are delivered to the Canonicalization Module (CM). Next, the CM maps the plain-text predicates to KG ontology and produces triples with their corresponding confidences. At this stage, the canonicalized triples are ready to be stored into the candidate-facts-triples-repository.
Afterwards, Knowledge Fusion module integrates triples which are stored in the CFTR and extracts triples with minimum confidence threshold predefined in the module. After this stage, triples are passed to human experts to check the correctness of the triples. The correct triples then are added to the FarsBase.

\subsection{Extraction Modules}
\subsubsection{PredPatt Extractor}
To present a multilingual state-of-the-art unsupervised baseline method alongside with our other extraction modules, we utilized PredPatt of Decomp project~\cite{white2016universal} on Universal Dependencies (UD) Project~\cite{nivre2016universal} for the Persian language. UD project provides a standard syntactic annotation which can be used in different languages. This project purposes unified grammatical representation which is valid cross-linguistically and includes, universal set of dependency relations and POS tags. Seraji et al. presented UD for the Persian language~\cite{seraji2016universal}.

While the UD project present syntactic annotation, Decomp project aims to propose a set of protocols for augmentation datasets in UD format with universal decompositional semantics. In more straightforward language, the UD project proposes syntactic annotation while Decomp project proposes semantic annotation in different languages. PredPatt is a software package for processing UD annotated corpora and annotating them with Decomp protocols. If we look at PredPatt as a black box, it takes a UD annotated dataset in the desired language as input, and its output is that dataset augmented with predicate-argument information alongside other syntactic and semantic annotations. Different studies propose that PredPatt can be utilized as an Open IE system~\cite{govindarajan2019decomposing,claro2019multilingual} and it overcomes other state-of-the-art Open IE systems~\cite{zhang2017evaluation}.  

Our main idea here is to utilize PredPatt to extract sets of predicate-arguments from input Persian sentences. For every input sentence, this module provides a predicate and two or more arguments. All possible permutation of arguments are then considered as subjects and objects, and the resulting triples are sent to the output as candidate relationships. For example, if three arguments are specified for a predicate in a sentence in the form of (arg1, pred, arg2, arg3), six triples are generated at the output including (arg1, pred, arg2), (arg1, pred, arg3), (arg2, pred, arg1), (arg2, pred, arg3), (arg3, pred, arg1), and (arg3, pred, arg2). It should be noted that these triples are not necessarily correct relationships. In the next section, we will discuss the details of the performance of this module as a state-of-the-art baseline on our gold dataset.

\subsubsection{Dependency Pattern Extractor}
The dependency pattern information extractor is the first of our novel information extractor modules. Extraction with dependency patterns is an innovative method that attempts to extract information triples using ``unique dependency trees''. 
The main idea is that sentences with the same structure of dependency trees can contain similar relationships. That means if a particular form of a dependency tree is observed in a sentence and the sentence contains a relationship, other sentences with the same dependency tree structure can probably contain a relationship with the same pattern.  Extraction with dependency pattern is based on the fact that if a sentence contains some triples, other sentences with the same structure (same dependency pattern) contain the same triples too. In such cases, the subject, object, and predicate can be extracted from the words with the same indexes in all sentences. For example, if in a sentence, the first word is the subject, then in all other sentences with the same dependency pattern, the subject can be extracted from the first word as well~\cite{asgari2019farsbase}.

Two dependency parse trees haves the same dependency pattern if similar trees are generated when words are replaced with their corresponding POS tags.  In other words, the sequences of the POS tags in both sentences are precisely the same, and also the dependency type and head of each word in the same position are equal.

To build this module, we utilized expert annotators to extract the desired rules. It should be noted that despite the use of human annotators, this method is not a supervised method. Because experts once extract the rules of this method and then can be used forever, and in fact, the system works as an unsupervised rule-based system which does not need any learning dataset.

In order to produce the rules, we first extracted dependency patterns for all of the sentences in a raw text corpus and then calculated most frequent dependency patterns in the corpus. We then asked human annotators to inspect the sentences of each of those patterns and annotate subject, object and predicate, for all of the relationships which can be mined from the sentences. In this way, we can determine which of the word positions in a given dependency tree and in what order, define the components of a relationship, namely the subject, object, and predicate. In this fashion, we were able to extract a set of relationship patterns in specific dependency tree structures. Currently, 3499 frequent dependency patterns are extracted automatically from Wikipedia texts, and experts annotate 6\% of them.

This module operates based on these extracted rules. For every new sentence the module receives, it first produces the dependency tree of the sentence and then compares this tree with the tree structure of the patterns generated earlier. If a matched dependency pattern is found, the module extracts a relation based on annotations defined for the pattern.

\subsubsection{\unsupervisedModule{} (\unsupervisedModulShorten{})}
In the \unsupervisedModule{}, the goal is to extract relation instances from an unlabeled and unannotated text, without any predefined relations as a train set. This method uses grammatical structures of sentences to extract relation instances. In this approach, all the predicates are based on verbs of the sentences, not in the predefined relations set, which is already known. Therefore, the evaluation of the results is not easy so that human experts intervention is required. Like other IE modules, another problem which was mentioned before is that predicates in the relation instances must be canonicalized to FarsBase ontology. 

This approach can be classified as an open information extraction approach. For the first time, KnowItAll \cite{etzioni2004web} introduced unsupervised knowledge extraction. The first model introduced in this project was TextRunner \cite{banko2007open}. TextRunner is comprised of three main components, namely Self-supervised learner, single-pass extractor, and redundancy-based assessor. 

Wikipedia-based open extractor(WOE) \cite{wu2010open} is another method of unsupervised knowledge extraction, which extracts knowledge from Wikipedia articles and uses dependency path between entities to improve the performance. OLLIE \cite{schmitz2012open} and BONIE \cite{saha2017bootstrapping} are two more recent unsupervised methods used for knowledge extraction.

Our system uses a different unsupervised method for triple extraction, i.e. dependency parsing and constituency tree are combined to extract the triples. Details of this module have been published in \cite{asgari2019farsbase}.

\subsubsection{RePersian - an automated ReVerb approach for the Persian language}
ReVerb \cite{etzioni2011open} is an unsupervised knowledge extraction method that was first introduced and applied in the KnowItAll \cite{etzioni2004web} project. ReVerb aimed to improve the performance of TextRunner by imposing lexical and syntactical constraints on relations. We employ RePesian~\cite{sahebnassagh2020repersian}, which uses a novel approach to produce regular expressions in ReVerb automatically. To find out the most frequent regular expressions, Dadegan Dependency Parsing Treebank~\cite{dadegan2012persian} processed with an automated algorithm. The algorithm found the best regular expressions for the subject, object, and Mosnad in the Persian language. We have already explained the details of this method in the previously published RePersian~\cite{sahebnassagh2020repersian} article.

\subsubsection{Distant Supervision}
One of the main challenges in relation extraction is the time and effort needed to make a manually labelled dataset. Distant supervision module utilizes a knowledge base to address this issue. To be more precise, the idea is that if a sentence contains a pair of entities which are relevant to each other in the knowledge base, this sentence may represent the required semantic relation. 
Mintz et al. \cite{mintz2009distant} used DBpedia and a collection of Wikipedia articles to make a distantly supervised dataset, and then extracted thousands of new relation instances from that dataset. There have been many modifications to this method, like using multi-instance learning approach proposed by Manning \cite{surdeanu2012multi}. The single-instance approach assumes that each sentence has one pair of entities and only one semantic relation between them, but multi-instance approach jointly models all the instances of a pair of entities in text and all their relations. Recent studies have employed word embedding and various type of deep neural networks to perform relation extraction from distantly supervised datasets. Using word embedding eliminates the need for manually extracted features from sentences.
Trisedya et al. \cite{distiawan2019neural} have recently proposed another distantly supervised relation extractor for knowledge base enrichment. Their neural end-to-end relation extraction model integrates the extraction and canonicalization tasks. Thus, their model reduces the error propagation between relation extraction and Named Entity Disambiguation. As a result, the existing approaches which are error-prone can be addressed by this model. To obtain high-quality training data, they adapt distant supervision and augment it with co-reference resolution and paraphrase detection.

We use FarsBase, as our knowledge base, and the unstructured text that our crawler module retrieves to create a Distantly Supervised dataset (DS-dataset). First, all the sentences of the raw text are checked by the module. Any sentence that contains a pair of related FarsBase entities is considered as a candidate containing the semantic relation of the same relation of entity pair. Then, a piecewise convolutional neural network (PCNN) with multi-instance learning is employed to extract knowledge from the dataset. Should be noted that our distantly supervised dataset is the first dataset provided for the Persian language. Moreover, the distant supervision component also produces frequent tokens and compound verbs for every predicate, which is used in our canonicalization and integration phases.

\subsubsection{TokensRegex}
\begin{figure}
	\frame{\includegraphics[width=\linewidth]{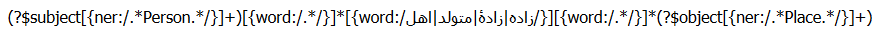}}
	\caption{A TokensRegex example matching fb:nationality predicate}
	\label{fig:TokensRegex}
\end{figure}

TokensRegex extraction module is a rule-base extraction module, which works based on Stanford TokensRegex \cite{chang2014tokensregex}. TokensRegex is a framework for specifying regular expressions over sequences of tokens and their additional side information, such as NER and POS tags, for identifying and acting on patterns in text. After the entity linking on the text, we have used the class of each entity instead of using named entity tags. Human experts have defined 166 TokensRegex rules for 58 frequent predicates on our gold dataset based on words, POS tags, and entity classes. Subject and object are defined in the body of each regular expression, and each expression refers to a predicate in FarsBase. Then, these rules were considered as the rules for relation extraction for this extractor module. Each sentence of the input raw text is examined according to the rules, and if the desired pattern is found, a relationship will be extracted. Figure \ref{fig:TokensRegex} shows an example of TokensRegex that matches a sentence, and Token Regex module extracts a fb:nationality triple.

\subsection{Relation to Knowledge Mapper}
\label{sec:mapper}
To extract a standard triple for knowledge graph, we must link two types of data:
\paragraph{Entity Linking (Relation arguments to KG entities)}
In KG, all subjects must be an entity. Our entity linker, ParsEL, finds all entity mentions in the tokens of the sentence and disambiguates and links them to the knowledge graph. Disambiguation process is a challenging task. Entity linker assigns a confidence value to each link. If all tokens of an argument belong to the same entity, the argument will be linked to it. When there are two different entities for an argument with different confidence values, two triples are generated using both entities with lower confidence values (weighted by entity confidences). These triples have the chance to be transferred to the final KG by Knowledge Fusion Module if they are extracted from multiple sources.
\paragraph{Canonicalization (Relation type to KG ontology predicate)}
We use a canonicalization module (CM) component to map relations extracted by four IE components to Knowledge Graph ontology. A baseline method proposed alongside with our FarsBase Canonicalization module.

\textbf{Baseline:} To implement a canonicalization baseline method, we examined the state-of-the-art methods as likely to be applicable to the Persian language. CESI~\cite{vashishth2018cesi} is one of the canonicalization methods introduced in the related works section. CESI needs side information to perform canonicalization, which is earned from AMIE~\cite{galarraga2013amie} and Stanford KBP~\cite{surdeanu2012multi}, latest is not present the Persian language. Therefore, this method could not be implemented for the Persian language. Lin and Chen approach for canonicalization is another state-of-the-art method, which we have considered to implement for the Persian language. Like CESI, this method utilizes the side information of the original data sources, including the entity linking knowledge, the types of the candidate entities detected, as well as the domain knowledge of the source text. The entity linking knowledge and domain knowledge in this approach is not available for the Persian language. Thus, this method could not be implemented for the Persian language as well. To prepare a baseline for canonicalization, we implemented the Galárraga and Heitz method~\cite{galarraga2014canonicalizing} for the Persian language. This method was the only method that could be implemented for the Persian language among the state-of-the-art methods introduced in the related works section.

For this aim, we first gave the DS-dataset generated by the distant supervision module to the PredPatt, Dependency Pattern, RePersian and \unsupervisedModulShorten{} modules and several tuples were constructed. Then we gave all these generated tuples, which were 3059530 relationships, to the AMIE algorithm, and a number of rules were extracted. Then, we cluster each pair of rules if they have this logical relationship: 

(s1, Pred1, o1) => (s1, Pred2, o1) AND (s2, Pred2, o2) => (s2, Pred1, o2)

Subsequently, a mapping table extracted from these clusters in the form of (raw text relations to the standard FarsBase relations), then this mapping table was controlled and filtered by human experts to remove the incorrect mappings, and finally, a mapping table was prepared as a baseline.

It should be noted here that in Galárraga and Heitz method~\cite{galarraga2014canonicalizing}, the input data is extensive, and therefore this method offers outstanding results in English. Nevertheless, there is no such big data in Persian. All extractors have been applied to the DS-dataset, and more than three million tuples extracted, which is far less from corresponding dataset in English. As a result, only about six thousand rules were extracted by the AMIE algorithm. Among these rules, a large number of them, extracted as transitive relations, which are not useful in canonicalization. Also, some extracted rules did not follow the aforementioned logical relationship. Eventually, 22 clusters of rules were obtained. Also, in some of these clusters, there were only rules between FarsBase predicates, and no plain text relationship was available in this type of cluster, and practically no new information was produced for the canonicalization. Some of these clusters only included plain text relationship, without any FarsBase predicate. In the latter case, we mapped these types of clusters to the FarsBase predicates manually. With all these modifications, there are still 15 clusters left to prepare the mapping table. As a result, this algorithm did not provide good results due to the lack of sufficient resources in Persian, despite the fact that it is the only available state-of-the-art method that could be implemented for the Persian language.

It should be noted that we did not use available Persian raw text corpora larger than Persian Wikipedia, such as MirasText~\cite{SABETI18.385}, to create DS-dataset. We created our  DS-dataset with the help of Persian Wikipedia containing more than 125 million words, while The MirasText corpus has more than 1.4 billion words. Despite its much larger size, the MirasText corpus is not suitable for use in this case. MirasText have been extracted from the texts of Persian news websites. To create a distantly supervised dataset in Persian, we need to look for entities of FarsBase, the only available knowledge base for the Persian, in a raw corpus. In contrary to the MirasText, it is expected that in the Persian Wikipedia sentences, there are many more sentences in which the FarsBase entities are presentو for two reasons: (I) Because most of the FarsBase triplets are extracted from Wikipedia infoboxes and (II) most of the entities in the infoboxes are included in the sentences of Wikipedia articles, while this condition exists with much less density in the MirasText.

\textbf{FarsBase Canonicalization module:} This module performs some steps to map a relation to standard FarsBase predicates which are explained below:
\begin{enumerate}
	\item If the extracted relation predicate phrase exists in the mapping set of the Wikipedia to FarsBase, the algorithm will map it to the corresponding KG predicate. The mapping table of the Wikipedia infobox predicates to FarsBase ontology had been introduced in our previous work \cite{asgari2019farsbase}.
	\item If the previous criterion is not satisfied, mapper uses the information extracted in Distant Supervision Component to find the best match of candidate KG predicate. Mapper matches words and compound verbs of the sentence of the predicate with corresponding information from distant supervision component. For each candidate predicate match, the mapper denotes one positive point for the candidate predicate. Finally, the mapper selects the predicate with the highest rank.
\end{enumerate}
\subsection{Knowledge Fusion Module}
Knowledge Fusion Module is a simple ensemble classifier. This module accepts a triple as a fact and sends it to the next component if one of these two conditions is satisfied:
\begin{enumerate}
	\item If a triple (with any confidence) is extracted by more than one (two to five) extractor.
	\item If a triple is extracted by just one extractor while its corresponding confidence is higher than a specified threshold. 
\end{enumerate}

\section{Experimental Results}
\label{experiments}
In this section, we introduce FarsBase-KBP Gold Dataset containing 22015 facts which are labelled and verified by human experts as gold data. This corpus has been used in the evaluation of all the components of FarsBase-KBP. Note that, all the modules are unsupervised, and the corpus is created only for the evaluation purposes. In this section, we introduce FarsBase-KBP Gold Dataset and then provide experimental results. 

\subsection{FarsBase-KBP Gold Dataset}
We trained FarsBase-KBP and its components over a corpus of 22015 facts which are labelled by human experts as gold data. To build this corpus, we searched for subject and object of each FarsBase triples in Wikipedia articles in the Persian language. More precisely, a sentence is a candidate for a triple's predicate if the sentence contains both subject and object of the considered triple. 
Also, we have chosen 406 distinct frequent FarsBase predicates for which the sentences are annotated by human experts. At last, we collected a gold dataset with 22015 sentence and its corresponding subject, object, and predicates.

This corpus has some automatic preprocessing as follows:
\begin{itemize}
	\item Tokens of each sentence.
	\item Part of speech tags (POS).
	\item Named Entity Recognition tags (NER).
	\item Dependency Parsing trees.
	\item Linked entities to the FarsBase.
	\item FarsBase classes for subject and object.
\end{itemize}

\begin{figure}
	\frame{\includegraphics[width=\linewidth]{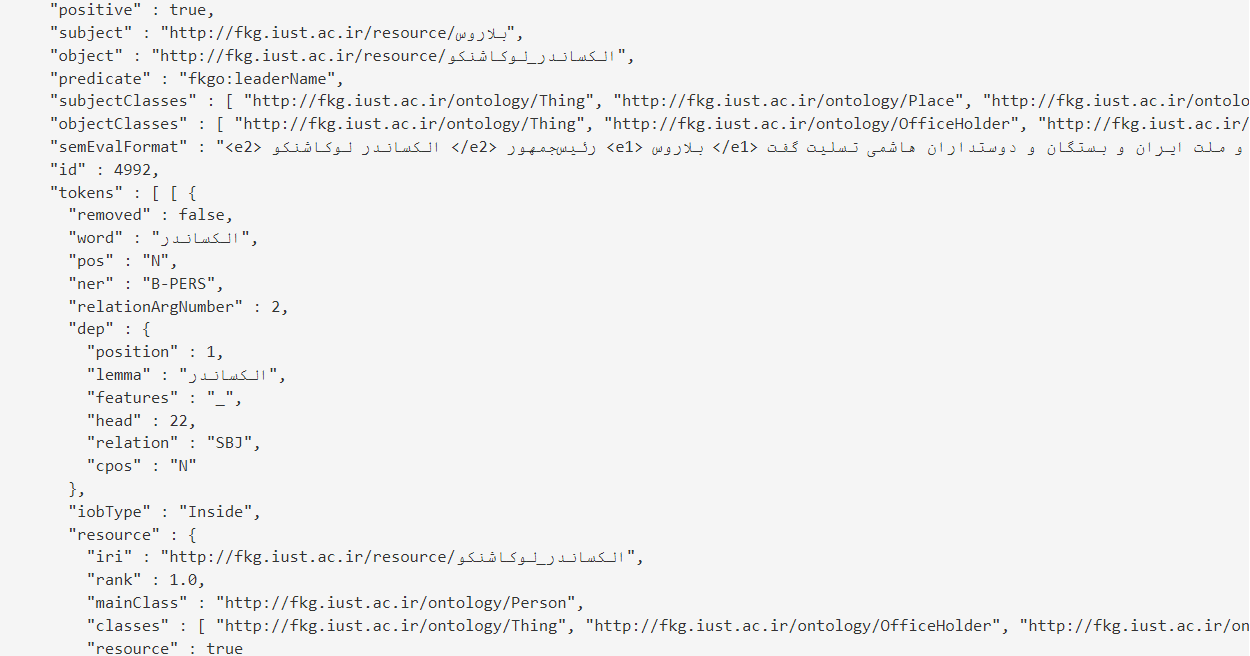}}
	\caption{An instance of gold corpus used to train FarsBase-KBP components}
	\label{fig:GoldData}
\end{figure}

Figure \ref{fig:GoldData} shows an instance of each triple in JSON format. In this example, the subject is ``Belarus'', and the object is ``Alexander Lukashenko'' and both are linked to the corresponding entities in FarsBase knowledge graph. The predicate is ``fkgo:leaderName'' which is the standard ontology-predicate in FarsBase and is approved by a human expert; thus this attribute is the gold part of the corpus. Other features like the subject, object, KG classes, tokens, NER tags, and POS tags are also included in the gold data.

\begin{figure}
	\frame{\includegraphics[width=\linewidth]{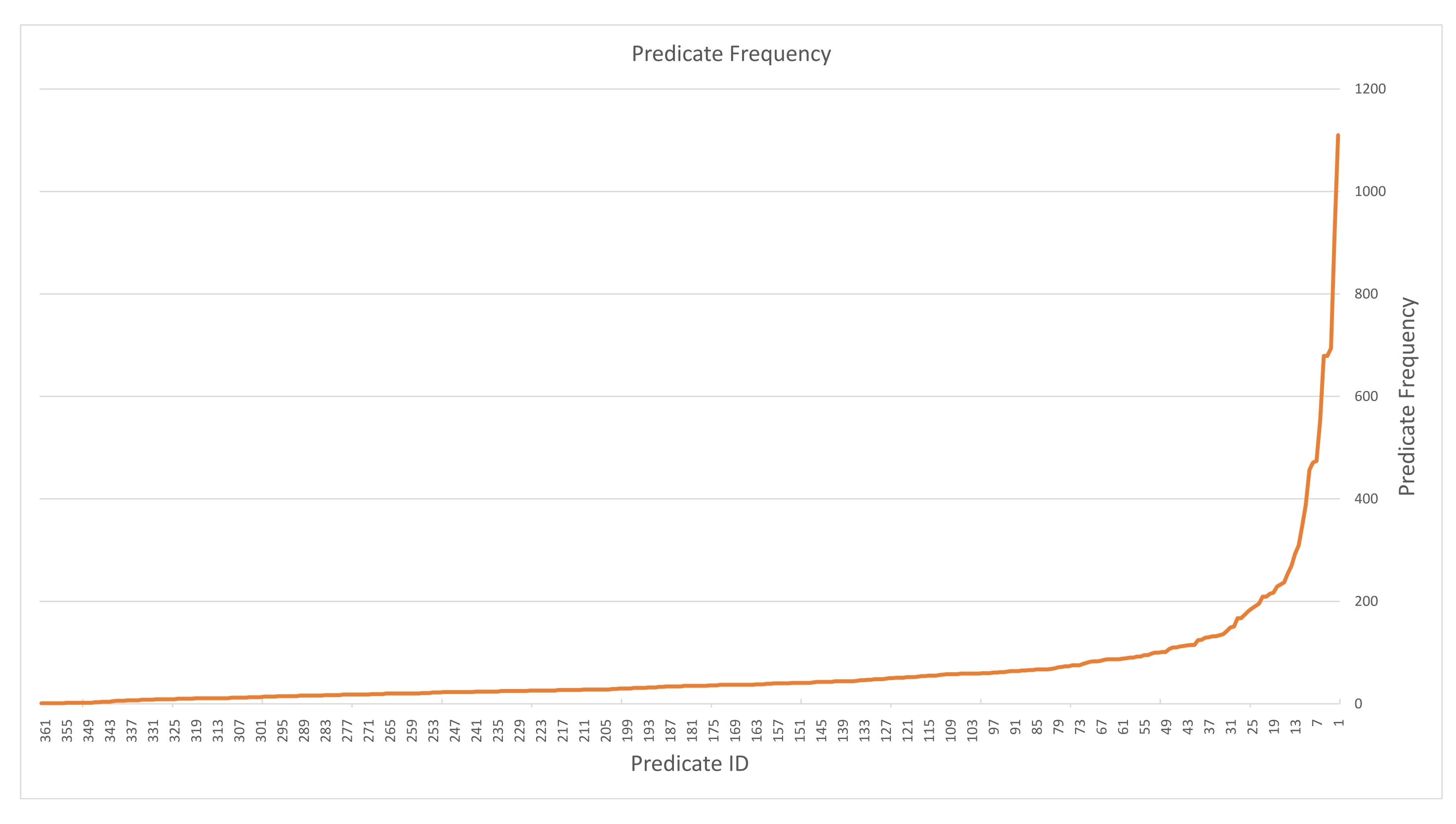}}
	\caption{Distribution of predicate instances based on frequency in the gold data}
	\label{fig:PredicateFrequency}
\end{figure}

Figure \ref{fig:PredicateFrequency} shows the distribution of predicate instances frequency in the gold data, which is sorted by frequency. Observation shows that there exists a normal distribution among them. In Table  \ref{tab:expcond}, an example of the most frequent predicates in the gold corpus has been written.

\begin{table}[htbp]
\centering
\caption{The most frequent predicates in the gold corpus}
\label{tab:expcond}
\begin{tabular}{lclc}
\hline
Predicate & Instance Count & Predicate & Instance Count\\
\hline
fkgo:location       & 1110  &  fkgo:type         & 457\\
foaf:name           & 915   &  fkgo:order        & 388\\
fkgo:birthPlace     & 694   &  fkgo:field        & 347\\
fkgo:occupation     & 679   &  fkgo:team         & 309\\
fkgo:speciality     & 679   &  fkgo:releaseDate  & 292\\
fkgo:genus          & 555   &  fkgo:language     & 269\\
fkgo:nationality    & 474   &  fkgo:family       & 255\\
fkgo:birthDate      & 471   &  fkgo:notableWork  & 237\\

\end{tabular}
\end{table}

\subsection{Evaluations}

Table \ref{tab:Eval} shows results and evaluation metrics of each FarsBase-KBP extraction module. We fed system with the previously mentioned corpus and used it as a gold dataset to evaluate the Precision, Recall, and F\textsubscript{1} (harmonic mean of precision and recall) of each module. Note that, every sentence in the dataset may have multiple triples, but one triple per sentence has been defined. Also, each module may extract more than one triple for some of the sentences. To calculate the recall, we only consider the gold triples of our dataset.
\begin{table}[htbp]
	\centering
	\caption{Evaluation Statistics}
	\label{tab:stat}
	%\tiny
	\begin{tabular}{llllll}
\hline
    Module   Name      & Triples        & Corrects      & Wrongs         & OSO            & Tri./Sen.       \\
\hline
    DependencyPattern  & 418            & 71            & 24             & 323            & 0.019           \\
    DistantSupervision & 17745          & \textbf{4632} & \textbf{13113} & 0              & 0.806           \\
    PredPatt           & \textbf{66384} & 13            & 82             & \textbf{66289} & \textbf{3.0154} \\
    RePersian          & 7865           & 51            & 241            & 7573           & 0.3573          \\
    TokensRegex         & 37351          & 3306          & 917            & 33128          & 1.6966          \\
    PSIE               & 44809          & 94            & 484            & 44231          & 2.0354         \\
\hline
	Fusion			   & 39375          & 3730          & 1280           & 34365          & 1.8119     \\
\hline
	\end{tabular}
\end{table}
\begin{table}[htbp]
	\centering
	\caption{Evaluation Results}
	\label{tab:Eval}
	%\tiny
	\begin{tabular}{llll}
\hline
	Module Name       &    Precision   &    Recall     &      F\textsubscript{1}       \\
\hline
    DependencyPattern  & 0.7474          & 0.0032          & 0.0064          \\
    DistantSupervision & 0.261           & \textbf{0.2104} & 0.233           \\
    PredPatt           & 0.1368          & 0.0006          & 0.0012          \\
    RePersian          & 0.1747          & 0.0023          & 0.0046          \\
    TokensRegex.json    & \textbf{0.7829} & 0.1502          & 0.252           \\
    PSIE               & 0.1626          & 0.0043          & 0.0083          \\
\hline
	Fusion			   & 0.7313          & 0.1779          & \textbf{0.2862}     \\
\hline
	\end{tabular}
\end{table}

The first five rows in the table shows the following information for the five relation and knowledge extraction modules:
\begin{itemize}
	\item Triples: The number of triples which are extracted from 22015 sentences.
	\item Corrects: The number of extracted triples which are correct and exists in the gold dataset.
	\item Wrongs: The number of extracted triples that their entities are correct but the relation is incorrect.
	\item OSO: The number of extracted triples not existing in the gold dataset.
	\item Tri. /Sen.: The number of extracted triples per sentence.
\end{itemize}

The last row shows these metrics for Fusion module when the confidence threshold is 0.9. As can be seen, the fusion module outperformed all of the individual modules based on f1 metric, while its precision and recall are comparable with the performance of the best extractors. The experts must approve each triple before adding to the knowledge graph. Therefore, the precision metric has the highest importance during the entire extraction process. Human intervention in a web-scale system, such as our proposed system, as well as other web-sclase machine reading systems, is for verification, not to create a supervised system. In this case, some human users are asked to review and accept or reject the triples which the system has calculated relatively high confidence for them. We can reduce the confidence threshold, and as a result, the recall will reach much higher results. However, this dramatically increases the number of triples that should be examined by human users and changes their function from a verifier to an annotator. Because human verifiers help the system voluntarily, providing such a volume of incorrect triplets for them to verify, will exhaust their motivation to continue doing so. In other words, we want human users to have only the role of verification and provide them with triples of high confidence to verify. Also, if for any reason human users are not available to verify triples, the system must work with the maximum precision to prevent the promotion of wrong triples to the FarsBase, even if it is at the cost of reducing the recall.  

\begin{figure}
	\frame{\includegraphics[width=\linewidth]{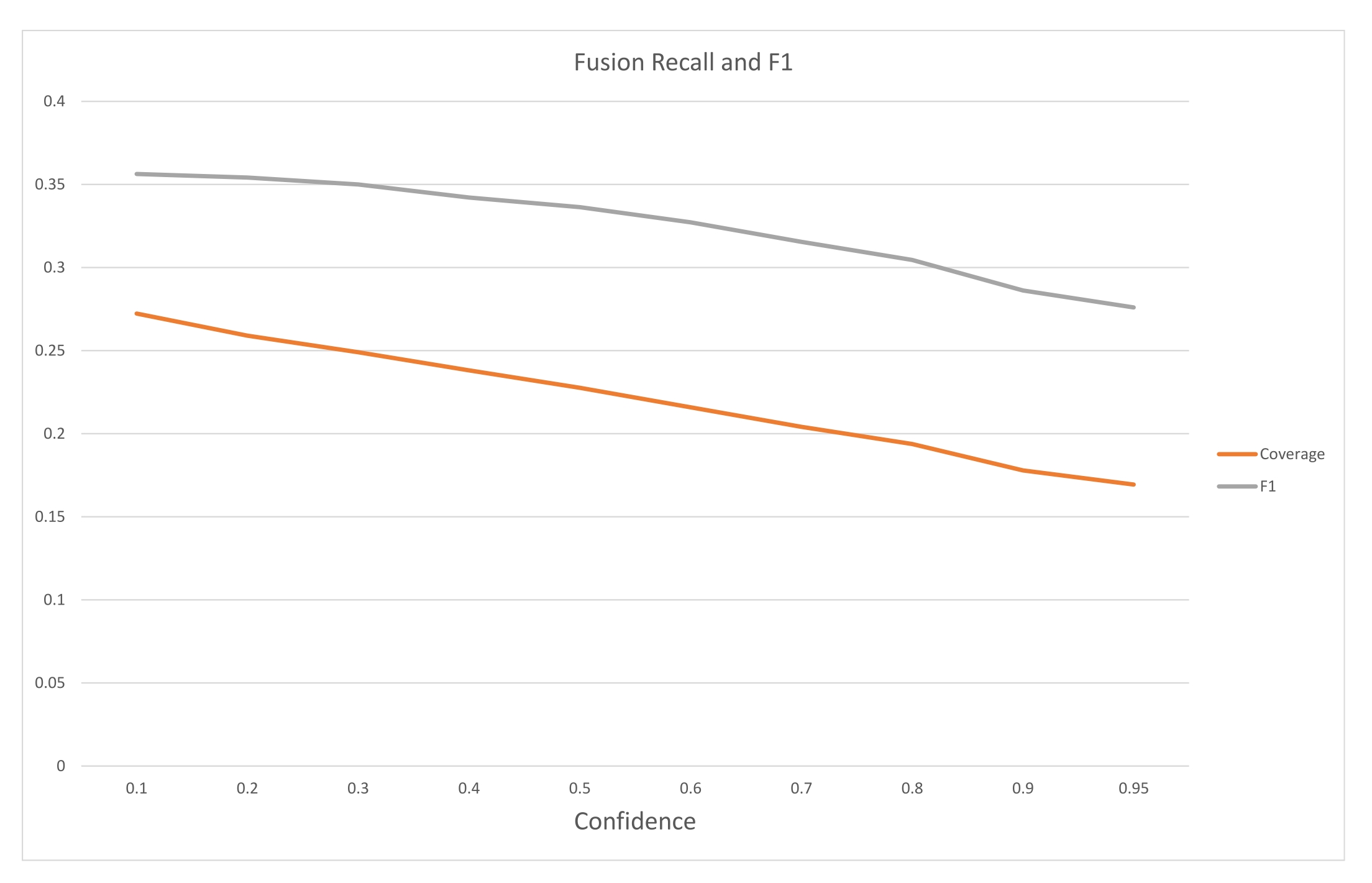}}
	\caption{Effect of changing the confidence threshold on Recall and F\textsubscript{1} measure of the Fusion module}
	\label{fig:Recall-F1-Threshold}
\end{figure}

Figure \ref{fig:Recall-F1-Threshold} shows the effect of changing the value of the confidence threshold on the Fusion module in contrast to recall and F\textsubscript{1} measures. The chart depicts values of recall and f-measure of fusion module for the different thresholds, and it shows the inverse relationship between the threshold and recall values. The plotted bullets define recall and  F\textsubscript{1} measures for the other modules. Distant supervision meets the recall line before threshold=0.1, but other modules meet the line after 0.9. F\textsubscript{1} is also decreasing from 0.1. Because of the low precision of Distant Supervision module, it meets the line after 0.4 with the F\textsubscript{1} line. Other modules meet the line after 0.9. As we expect, the higher thresholds lead to a lower recall. 
%Decreasing F\textsubscript{1} indicates that the downward slope of recall is greater than the upward slope of precision; thus, we prefer precision to recall. \majid{we must discuss about the last sentence.}

\begin{figure}
	\frame{\includegraphics[width=\linewidth]{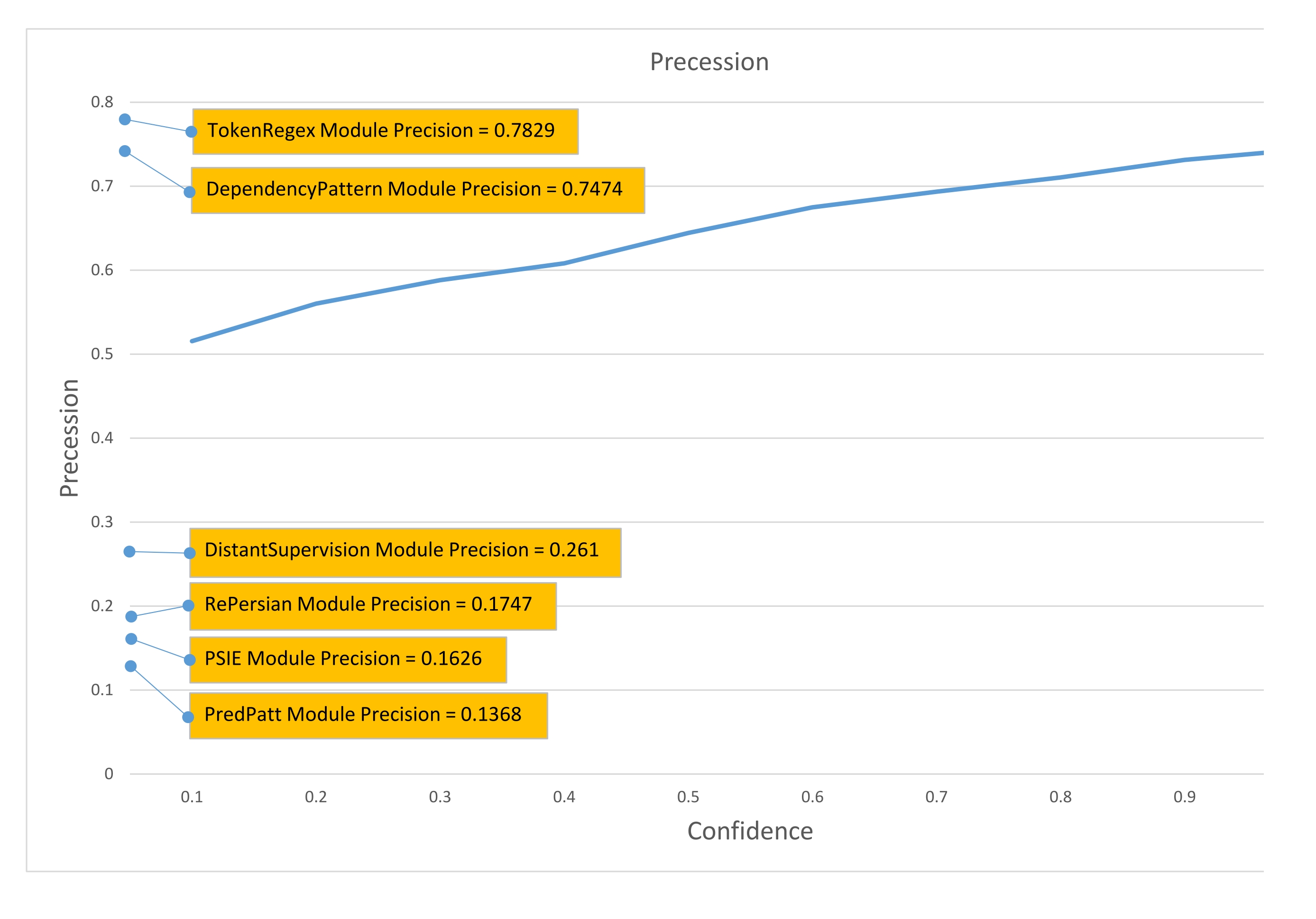}}
	\caption{Effect of changing the confidence threshold on Precision of the Fusion module}
	\label{fig:threshold_changes}
\end{figure}

Figure \ref{fig:threshold_changes} shows the effect of changing the confidence threshold on the precision value of the Fusion module. As we expect, higher thresholds lead to higher precision. The bullets on the chart show the precision of each single extraction module. According to the results, the precision metric of the fusion module outperforms two of the extraction modules when threshold=0, and when threshold=0.1, the fusion module extracts triples with precision higher than all of the modules except DependencyPattern module. Finally, the fusion module overtakes all the extraction modules when threshold=0.8.

Comparing \ref{fig:Recall-F1-Threshold} and Figure \ref{fig:threshold_changes}, we can induce that while higher threshold results in higher precision, it has a converse effect on recall and F\textsubscript{1} measures.

Table \ref{tab:Common} shows the number of commonly extracted triples between each pair of modules. The second column shows the number of all triples extracted by the modules, and the third column illustrates the number of corrected triples, which are existed in the gold data. Note that, the common triples are calculated on all triples extracted by modules, not only the triples which are existed in the gold data. The Acquired results describe that Distant Supervision and TokensRegex modules can extract the most common triples (2300). To find out which pair of modules extract similar triples, we should pay attention to the number of triples extracted by the modules, and consider the portion of common triples to the average number of triples extracted by both modules. 

\begin{table}[htbp]
	\centering
	\caption{The number of common extracted triples between every pair of modules}
	\label{tab:Common}
	\begin{tabular}{lllllll}
\hline
    Module   & DisSup & TokReg & DepPat & PSIE & RePer & PredPatt \\
\hline    
    DisSup   & 0      & 2300   & 104    & 36   & 104   & 20       \\
    TokReg   & 2300   & 0      & 36     & 124  & 334   & 8        \\
    DepPat   & 104    & 36     & 0      & 20   & 36    & 10       \\
    PSIE     & 36     & 124    & 20     & 0    & 442   & 62       \\
    RePer    & 104    & 334    & 36     & 442  & 0     & 496      \\
    PredPatt & 20     & 8      & 10     & 62   & 496   & 0       \\
\hline
    Triples  & 17745  & 37351  & 418    & 44809 & 7865  & 66384    \\
    Corrects & 4632   & 3306   & 71     & 94    & 51    & 13      \\
\hline
	\end{tabular}
\end{table}

\section{Conclusion}
\label{conclusions}
In this paper, we provide a novel gold corpus for training and evaluating knowledge extractors in the Persian language. We also introduce multiple unsupervised components for relation and knowledge extraction in the Persian language. The acquired results show that using a fusion module can increase the precision of knowledge extraction when it works above all individual extractor components. 

In future work, we will address the low recall and precision of some of our extractor components. We also will provide some new supervised and unsupervised modules to increase the recall and precision of the system. Also, we plan to add some extractors which will work on other types of resources, such as tabular data. Another improvement is to add a rule learner module to conclude facts and generate new triples. Also, by adding a type checker to the Knowledge Fusion module, the wrong triples based on the domain and range of each predicate can be filtered.
Moreover, one of the problems is when the modules extract many triples at the end of each day so that verifying all the triples will be a tough task. One of the potential solutions is to utilize crowdsourcing to accelerate the triple verification process. The output of PredPatt module is hugely affected by the quality of universal dependency parsing. Based on our observations, the output of current universal dependency parser for the Persian language has many errors, especially for long sentences. We can create another UD corpus by converting Dadegan dependency parser dataset and implement a better UD parser in the future works.

\section*{Acknowledgment}

The authors would like to thank Mr. Mehrdad Nasser and Ms. Raana Saheb-Nassagh at Data Mining Laboratory, Faculty of Computer Science, Iran University of Science and Technology, for implementing their methods, namely Distant Supervision RE and RePersian, for FarsBase. The authors also wish to express their sincere appreciation to Dr. Sayyed Ali Hossayni for his constructive guidance and valuable comments.
\section*{References}

%\begin{thebibliography}{5}
\bibliography{FarsBaseKBP}

\begin{thebibliography}{10}
\expandafter\ifx\csname url\endcsname\relax
  \def\url#1{\texttt{#1}}\fi
\expandafter\ifx\csname urlprefix\endcsname\relax\def\urlprefix{URL }\fi
\expandafter\ifx\csname href\endcsname\relax
  \def\href#1#2{#2} \def\path#1{#1}\fi

\bibitem{asgari2019farsbase}
M.~Asgari-Bidhendi, A.~Hadian, B.~Minaei-Bidgoli, Farsbase: The persian
  knowledge graph, Semantic Web 10~(6) (2019) 1169--1196.

\bibitem{auer2007dbpedia}
S.~Auer, C.~Bizer, G.~Kobilarov, J.~Lehmann, R.~Cyganiak, Z.~Ives, Dbpedia: A
  nucleus for a web of open data, in: The semantic web, Springer, 2007, pp.
  722--735.

\bibitem{navigli2010babelnet}
R.~Navigli, S.~P. Ponzetto, Babelnet: Building a very large multilingual
  semantic network, in: Proceedings of the 48th annual meeting of the
  association for computational linguistics, Association for Computational
  Linguistics, 2010, pp. 216--225.

\bibitem{vrandevcic2014wikidata}
D.~Vrande{\v{c}}i{\'c}, M.~Kr{\"o}tzsch, Wikidata: a free collaborative
  knowledge base, Communications of the ACM 57~(10) (2014) 78--85.

\bibitem{asgaribidhendi2020parsel}
M.~Asgari-Bidhendi, F.~Fakhrian, B.~Minaei-Bidgoli, Parsel 1.0: Unsupervised
  entity linking in persian social media texts, submited Manuscript to Computer
  Speech \& Language (2020).
\newblock \href {http://arxiv.org/abs/2004.10816} {\path{arXiv:2004.10816}}.

\bibitem{glass2018dataset}
M.~Glass, A.~Gliozzo, A dataset for web-scale knowledge base population, in:
  European Semantic Web Conference, Springer, 2018, pp. 256--271.

\bibitem{bollacker2008freebase}
K.~Bollacker, C.~Evans, P.~Paritosh, T.~Sturge, J.~Taylor, Freebase: a
  collaboratively created graph database for structuring human knowledge, in:
  Proceedings of the 2008 ACM SIGMOD international conference on Management of
  data, AcM, 2008, pp. 1247--1250.

\bibitem{min2013distant}
B.~Min, R.~Grishman, L.~Wan, C.~Wang, D.~Gondek, Distant supervision for
  relation extraction with an incomplete knowledge base, in: Proceedings of the
  2013 Conference of the North American Chapter of the Association for
  Computational Linguistics: Human Language Technologies, 2013, pp. 777--782.

\bibitem{west2014knowledge}
R.~West, E.~Gabrilovich, K.~Murphy, S.~Sun, R.~Gupta, D.~Lin, Knowledge base
  completion via search-based question answering, in: Proceedings of the 23rd
  international conference on World wide web, ACM, 2014, pp. 515--526.

\bibitem{adel2018deep}
H.~Adel, Deep learning methods for knowledge base population, Ph.D. thesis,
  Ludwig Maximilian University of Munich, Germany (2018).

\bibitem{getman2018laying}
J.~Getman, J.~Ellis, S.~Strassel, Z.~Song, J.~Tracey, Laying the groundwork for
  knowledge base population: Nine years of linguistic resources for {TAC}
  {KBP}, in: Proceedings of the Eleventh International Conference on Language
  Resources and Evaluation ({LREC}-2018), European Languages Resources
  Association (ELRA), Miyazaki, Japan, 2018, pp. 1552--1558.

\bibitem{gangemi2017semantic}
A.~Gangemi, V.~Presutti, D.~Reforgiato~Recupero, A.~G. Nuzzolese, F.~Draicchio,
  M.~Mongiov{\`\i}, Semantic web machine reading with fred, Semantic Web 8~(6)
  (2017) 873--893.

\bibitem{consoli2015using}
S.~Consoli, D.~R. Recupero, Using fred for named entity resolution, linking and
  typing for knowledge base population, in: Semantic Web Evaluation Challenges,
  Springer, 2015, pp. 40--50.

\bibitem{dong2014knowledge}
X.~Dong, E.~Gabrilovich, G.~Heitz, W.~Horn, N.~Lao, K.~Murphy, T.~Strohmann,
  S.~Sun, W.~Zhang, Knowledge vault: A web-scale approach to probabilistic
  knowledge fusion, in: Proceedings of the 20th ACM SIGKDD international
  conference on Knowledge discovery and data mining, ACM, 2014, pp. 601--610.

\bibitem{suchanek2007yago}
F.~M. Suchanek, G.~Kasneci, G.~Weikum, Yago: a core of semantic knowledge, in:
  Proceedings of the 16th international conference on World Wide Web, ACM,
  2007, pp. 697--706.

\bibitem{hoffart2013yago2}
J.~Hoffart, F.~M. Suchanek, K.~Berberich, G.~Weikum, Yago2: A spatially and
  temporally enhanced knowledge base from wikipedia, Artificial Intelligence
  194 (2013) 28--61.

\bibitem{fader2011identifying}
A.~Fader, S.~Soderland, O.~Etzioni, Identifying relations for open information
  extraction, in: Proceedings of the conference on empirical methods in natural
  language processing, Association for Computational Linguistics, 2011, pp.
  1535--1545.

\bibitem{fan2010prismatic}
J.~Fan, D.~Ferrucci, D.~Gondek, A.~Kalyanpur, Prismatic: Inducing knowledge
  from a large scale lexicalized relation resource, in: Proceedings of the
  NAACL HLT 2010 first international workshop on formalisms and methodology for
  learning by reading, Association for Computational Linguistics, 2010, pp.
  122--127.

\bibitem{schmitz2012open}
M.~Schmitz, R.~Bart, S.~Soderland, O.~Etzioni, et~al., Open language learning
  for information extraction, in: Proceedings of the 2012 Joint Conference on
  Empirical Methods in Natural Language Processing and Computational Natural
  Language Learning, Association for Computational Linguistics, 2012, pp.
  523--534.

\bibitem{gashteovski2017minie}
K.~Gashteovski, R.~Gemulla, L.~Del~Corro, Minie: minimizing facts in open
  information extraction, in: Proceedings of the 2017 Conference on Empirical
  Methods in Natural Language Processing, 2017, pp. 2630--2640.

\bibitem{cetto2018graphene}
M.~Cetto, C.~Niklaus, A.~Freitas, S.~Handschuh, Graphene: Semantically-linked
  propositions in open information extraction, in: Proceedings of the 27th
  International Conference on Computational Linguistics, {COLING} 2018, Santa
  Fe, New Mexico, USA, August 20-26, 2018, Association for Computational
  Linguistics, 2018, pp. 2300--2311.

\bibitem{wu2012probase}
W.~Wu, H.~Li, H.~Wang, K.~Q. Zhu, Probase: A probabilistic taxonomy for text
  understanding, in: Proceedings of the 2012 ACM SIGMOD International
  Conference on Management of Data, ACM, 2012, pp. 481--492.

\bibitem{nakashole2011scalable}
N.~Nakashole, M.~Theobald, G.~Weikum, Scalable knowledge harvesting with high
  precision and high recall, in: Proceedings of the fourth ACM international
  conference on Web search and data mining, ACM, 2011, pp. 227--236.

\bibitem{niu2012deepdive}
F.~Niu, C.~Zhang, C.~R{\'{e}}, J.~W. Shavlik, Deepdive: Web-scale
  knowledge-base construction using statistical learning and inference, in:
  Proceedings of the Second International Workshop on Searching and Integrating
  New Web Data Sources, Istanbul, Turkey, August 31, 2012, Vol. 884 of {CEUR}
  Workshop Proceedings, CEUR-WS.org, 2012, pp. 25--28.

\bibitem{carlson2010toward}
A.~Carlson, J.~Betteridge, B.~Kisiel, B.~Settles, E.~R. Hruschka, T.~M.
  Mitchell, Toward an architecture for never-ending language learning, in:
  Twenty-Fourth AAAI Conference on Artificial Intelligence, 2010, pp.
  1306--1313.

\bibitem{mitchell2018never}
T.~Mitchell, W.~Cohen, E.~Hruschka, P.~Talukdar, B.~Yang, J.~Betteridge,
  A.~Carlson, B.~Dalvi, M.~Gardner, B.~Kisiel, et~al., Never-ending learning,
  Communications of the ACM 61~(5) (2018) 103--115.

\bibitem{moro2014entity}
A.~Moro, A.~Raganato, R.~Navigli, Entity linking meets word sense
  disambiguation: a unified approach, Transactions of the Association for
  Computational Linguistics 2 (2014) 231--244.

\bibitem{shi2019brief}
Y.~Shi, Y.~Xiao, L.~Niu, A brief survey of relation extraction based on distant
  supervision, in: International Conference on Computational Science, Springer,
  2019, pp. 293--303.

\bibitem{roy2019supervising}
A.~Roy, Y.~Park, T.~Lee, S.~Pan, Supervising unsupervised open information
  extraction models, in: Proceedings of the 2019 Conference on Empirical
  Methods in Natural Language Processing and the 9th International Joint
  Conference on Natural Language Processing (EMNLP-IJCNLP), 2019, pp. 728--737.

\bibitem{trisedya2019neural}
B.~D. Trisedya, G.~Weikum, J.~Qi, R.~Zhang, Neural relation extraction for
  knowledge base enrichment, in: 57th Annual Meeting of the Association for
  Computational Linguistics, Association for Computational Linguistics, 2019,
  pp. 229--240.

\bibitem{gao2019neural}
T.~Gao, X.~Han, R.~Xie, Z.~Liu, F.~Lin, L.~Lin, M.~Sun, Neural snowball for
  few-shot relation learning, CoRR abs/1908.11007.
\newblock \href {http://arxiv.org/abs/1908.11007} {\path{arXiv:1908.11007}}.

\bibitem{smirnova2018relation}
A.~Smirnova, P.~Cudr{\'e}-Mauroux, Relation extraction using distant
  supervision: A survey, ACM Computing Surveys (CSUR) 51~(5) (2018) 1--35.

\bibitem{putri2019aligning}
R.~A. Putri, G.~Hong, S.-H. Myaeng, Aligning open ie relations and kb relations
  using a siamese network based on word embedding, in: Proceedings of the 13th
  International Conference on Computational Semantics-Long Papers, 2019, pp.
  142--153.

\bibitem{vashishth2018cesi}
S.~Vashishth, P.~Jain, P.~Talukdar, Cesi: Canonicalizing open knowledge bases
  using embeddings and side information, in: Proceedings of the 2018 World Wide
  Web Conference, 2018, pp. 1317--1327.

\bibitem{galarraga2013amie}
L.~A. Gal{\'a}rraga, C.~Teflioudi, K.~Hose, F.~Suchanek, Amie: association rule
  mining under incomplete evidence in ontological knowledge bases, in:
  Proceedings of the 22nd international conference on World Wide Web, 2013, pp.
  413--422.

\bibitem{surdeanu2012multi}
M.~Surdeanu, J.~Tibshirani, R.~Nallapati, C.~D. Manning, Multi-instance
  multi-label learning for relation extraction, in: Proceedings of the 2012
  joint conference on empirical methods in natural language processing and
  computational natural language learning, Association for Computational
  Linguistics, 2012, pp. 455--465.

\bibitem{lin2019canonicalization}
X.~Lin, L.~Chen, Canonicalization of open knowledge bases with side information
  from the source text, in: 2019 IEEE 35th International Conference on Data
  Engineering (ICDE), IEEE, 2019, pp. 950--961.

\bibitem{galarraga2014canonicalizing}
L.~Gal{\'a}rraga, G.~Heitz, K.~Murphy, F.~M. Suchanek, Canonicalizing open
  knowledge bases, in: Proceedings of the 23rd acm international conference on
  conference on information and knowledge management, 2014, pp. 1679--1688.

\bibitem{mesquita2019knowledgenet}
F.~Mesquita, M.~Cannaviccio, J.~Schmidek, P.~Mirza, D.~Barbosa, Knowledgenet: A
  benchmark dataset for knowledge base population, in: Proceedings of the 2019
  Conference on Empirical Methods in Natural Language Processing and the 9th
  International Joint Conference on Natural Language Processing (EMNLP-IJCNLP),
  2019, pp. 749--758.

\bibitem{straka2015parsing}
M.~Straka, J.~Hajic, J.~Strakov{\'a}, J.~Hajic~Jr, Parsing universal dependency
  treebanks using neural networks and search-based oracle, in: International
  workshop on treebanks and linguistic theories (tlt14), 2015, pp. 208--220.

\bibitem{taghizadeh2018cross}
N.~Taghizadeh, H.~Faili, J.~Maleki, Cross-language learning for arabic relation
  extraction, in: Fourth International Conference On Arabic Computational
  Linguistics, {ACLING} 2018, November 17-19, 2018, Dubai, United Arab
  Emirates, Vol. 142 of Procedia Computer Science, Elsevier, 2018, pp.
  190--197.

\bibitem{zakria2019relation}
G.~Zakria, M.~Farouk, K.~Fathy, M.~N. Makar, Relation extraction from arabic
  wikipedia, Indian Journal of Science and Technology 12 (2019) 46.

\bibitem{alarfaj2019towards}
A.~AlArfaj, Towards relation extraction from arabic text: a review,
  International Robotics \& Automation Journal 5~(5) (2019) 212--215.

\bibitem{sarhan2016semi}
I.~Sarhan, Y.~El-Sonbaty, M.~A. El-Nasr, Semi-supervised pattern based
  algorithm for arabic relation extraction, in: 2016 IEEE 28th International
  Conference on Tools with Artificial Intelligence (ICTAI), IEEE, 2016, pp.
  177--183.

\bibitem{peters2018deep}
M.~E. Peters, M.~Neumann, M.~Iyyer, M.~Gardner, C.~Clark, K.~Lee,
  L.~Zettlemoyer, Deep contextualized word representations, in: Proceedings of
  the 2018 Conference of the North American Chapter of the Association for
  Computational Linguistics: Human Language Technologies, {NAACL-HLT} 2018, New
  Orleans, Louisiana, USA, June 1-6, 2018, Volume 1 (Long Papers), Association
  for Computational Linguistics, 2018, pp. 2227--2237.

\bibitem{sun2019improved}
Y.~Sun, L.~Wang, C.~Chen, T.~Xia, X.~Zhao, Improved distant supervised model in
  tibetan relation extraction using elmo and attention, IEEE Access 7 (2019)
  173054--173062.

\bibitem{peng2017jointly}
N.~Peng, et~al., Jointly learning representations for low-resource information
  extraction, Ph.D. thesis, Johns Hopkins University (2017).

\bibitem{moradi2015commonsense}
M.~Moradi, B.~Vazirnezhad, B.~Mohammd, Commonsense knowledge extraction for
  persian language: A combinatory approach, Iranian Journal of Information
  Processing and Management 31~(1) (2015) 109--124.

\bibitem{shamsfard2004learning}
M.~Shamsfard, A.~A. Barforoush, Learning ontologies from natural language
  texts, International journal of human-computer studies 60~(1) (2004) 17--63.

\bibitem{momtazi2019statistical}
S.~Momtazi, O.~Moradiannasab, A statistical approach to knowledge discovery:
  Bootstrap analysis of language models for knowledge base population from
  unstructured text, Scientia Iranica 26 (2019) 26--39.

\bibitem{white2016universal}
A.~S. White, D.~Reisinger, K.~Sakaguchi, T.~Vieira, S.~Zhang, R.~Rudinger,
  K.~Rawlins, B.~Van~Durme, Universal decompositional semantics on universal
  dependencies, in: Proceedings of the 2016 Conference on Empirical Methods in
  Natural Language Processing, 2016, pp. 1713--1723.

\bibitem{nivre2016universal}
J.~Nivre, M.-C. De~Marneffe, F.~Ginter, Y.~Goldberg, J.~Hajic, C.~D. Manning,
  R.~McDonald, S.~Petrov, S.~Pyysalo, N.~Silveira, et~al., Universal
  dependencies v1: A multilingual treebank collection, in: Proceedings of the
  Tenth International Conference on Language Resources and Evaluation
  (LREC'16), 2016, pp. 1659--1666.

\bibitem{seraji2016universal}
M.~Seraji, F.~Ginter, J.~Nivre, Universal dependencies for persian, in:
  Proceedings of the Tenth International Conference on Language Resources and
  Evaluation (LREC'16), 2016, pp. 2361--2365.

\bibitem{govindarajan2019decomposing}
V.~Govindarajan, B.~V. Durme, A.~S. White, Decomposing generalization: Models
  of generic, habitual, and episodic statements, Transactions of the
  Association for Computational Linguistics 7 (2019) 501--517.

\bibitem{claro2019multilingual}
D.~B. Claro, M.~Souza, C.~Castell{\~a}~Xavier, L.~Oliveira, Multilingual open
  information extraction: Challenges and opportunities, Information 10~(7)
  (2019) 228.

\bibitem{zhang2017evaluation}
S.~Zhang, R.~Rudinger, B.~Van~Durme, An evaluation of predpatt and open ie via
  stage 1 semantic role labeling, in: IWCS 2017—12th International Conference
  on Computational Semantics—Short papers, 2017.

\bibitem{etzioni2004web}
O.~Etzioni, M.~Cafarella, D.~Downey, S.~Kok, A.-M. Popescu, T.~Shaked,
  S.~Soderland, D.~S. Weld, A.~Yates, Web-scale information extraction in
  knowitall:(preliminary results), in: Proceedings of the 13th international
  conference on World Wide Web, ACM, 2004, pp. 100--110.

\bibitem{banko2007open}
M.~Banko, M.~J. Cafarella, S.~Soderland, M.~Broadhead, O.~Etzioni, Open
  information extraction from the web, in: {IJCAI} 2007, Proceedings of the
  20th International Joint Conference on Artificial Intelligence, Hyderabad,
  India, January 6-12, 2007, 2007, pp. 2670--2676.

\bibitem{wu2010open}
F.~Wu, D.~S. Weld, Open information extraction using wikipedia, in: Proceedings
  of the 48th annual meeting of the association for computational linguistics,
  Association for Computational Linguistics, 2010, pp. 118--127.

\bibitem{saha2017bootstrapping}
S.~Saha, H.~Pal, et~al., Bootstrapping for numerical open ie, in: Proceedings
  of the 55th Annual Meeting of the Association for Computational Linguistics
  (Volume 2: Short Papers), 2017, pp. 317--323.

\bibitem{etzioni2011open}
O.~Etzioni, A.~Fader, J.~Christensen, S.~Soderland, et~al., Open information
  extraction: The second generation, in: Twenty-Second International Joint
  Conference on Artificial Intelligence, 2011, pp. 3--10.

\bibitem{sahebnassagh2020repersian}
R.~Saheb-Nassagh, M.~Asgari-Bidhendi, B.~Minaei-Bidgoli, Repersian - an
  efficient open information extraction tool in persian, in: 2020 6th
  International Conference on Web Research (ICWR), IEEE, (accepted) 2020, p. to
  appear.

\bibitem{dadegan2012persian}
D.~R. Group, et~al., Persian dependency treebank version 0.1 (2012).

\bibitem{mintz2009distant}
M.~Mintz, S.~Bills, R.~Snow, D.~Jurafsky, Distant supervision for relation
  extraction without labeled data, in: Proceedings of the Joint Conference of
  the 47th Annual Meeting of the ACL and the 4th International Joint Conference
  on Natural Language Processing of the AFNLP: Volume 2-Volume 2, Association
  for Computational Linguistics, 2009, pp. 1003--1011.

\bibitem{distiawan2019neural}
B.~Distiawan, G.~Weikum, J.~Qi, R.~Zhang, Neural relation extraction for
  knowledge base enrichment, in: Proceedings of the 57th Conference of the
  Association for Computational Linguistics, 2019, pp. 229--240.

\bibitem{chang2014tokensregex}
A.~X. Chang, C.~D. Manning, Tokensregex: Defining cascaded regular expressions
  over tokens, Stanford University Computer Science Technical Reports. CSTR 2.

\bibitem{SABETI18.385}
A.~J. C. S. h. e. M.~N. Behnam~Sabeti, Hossein Abedi~Firouzjaee, A.~Vaheb,
  Mirastext: An automatically generated text corpus for persian, in:
  Proceedings of the Eleventh International Conference on Language Resources
  and Evaluation (LREC 2018), European Language Resources Association (ELRA),
  Paris, France, 2018, pp. 1174--1177.

\end{thebibliography}
%\end{thebibliography}

\end{document}